# Federated Learning based Latent Factorization of Tensors for Privacy-Preserving QoS Prediction

Shuai Zhong
The College of Computer and Information Science
Southwest University
Chongqing China
zhongzhiming@email.swu.edu.cn

Zengtong Tang
The College of Computer and Information Science
Southwest University
Chongqing China
xndx029311@email.swu.edu.cn

Di Wu
The College of Computer and Information Science
Southwest University
Chongqing China
wudi.cigit@gmail.com

*Abstract*—In applications related to big data and service computing, dynamic connections tend to be encountered, especially the dynamic data of user-perspective quality-of-service (QoS) in Web services. They are transformed into high-dimensional and incomplete (HDI) tensors which include abundant temporal pattern information. Latent factorization of tensors (LFT) is an extremely efficient and typical approach for extracting such patterns from an HDI tensor. However, current LFT models require the QoS data to be maintained in a central place (e.g., a central server), which is impossible for increasingly privacy-sensitive users. To address this problem, this article creatively designs a federated learning based on latent factorization of tensors (FL-LFT). It builds a data-density-oriented federated learning model to enable isolated users to collaboratively train a global LFT model while protecting users' privacy. Extensive experiments on a QoS dataset collected from the real world verify that FL-LFT shows a remarkable increase in prediction accuracy when compared to state-of-the-art federated learning (FL) approaches.

INTRODUCTION

Amid the swift advancement of cloud computing [1-4], various service providers supply an enormous variety of Web services that are functionally equivalent. Hence, determining the most appropriate service from an extensive list of candidates evolves into a tricky yet significant issue [5-7]. For users to determine whether a Web service is appropriate to their particular requirements, data is essential in that a Web service's quality-of-service (QoS) and previous QoS can be utilized to evaluate it. Consequently, QoS-based methods to select services have been explored [8-10] for over a decade.

Typically, user-perspective QoS data is acquired by the invocation of actual services [11-13]. However, some obstacles make the invocation interactions between a user group and a service group incomplete[14, 15]. Consequently, Qos data typically exhibits high dimensionality [16-19], severe sparsity [20-23], and time-varying dynamics [5, 24-26]. Nonetheless, as user-perspective QoS data is critical for the efficient selection of Web service, it becomes extremely crucial to accurately extract the temporal characteristics hidden within them [8, 27, 28].

Numerous advanced strategies have been proposed to tackle this matter, and latent factorization of tensors (LFT)-based ones are effective and typical [8, 14, 29]. It represents user-service-time interactions as high-dimensional and incomplete (HDI) tensors and derives the low-rank approximation to the resulting HDI tensor through the tensor decomposition principle [16, 27, 30, 31]. When utilizing tensors to represent QoS data, it naturally considers dynamic characteristics [32]. Besides, by utilizing the density-oriented modeling method only to focus on observed data of the target tensor, it has the benefit of excellent computational and storage efficiency [14, 33, 34].

Currently, people are increasingly worried about privacy problems [35-38]. The General Data Protection Regulation [39] forbids businesses from collecting, distributing, or utilizing user data without the user's explicit permission. Additionally, data owners (e.g., data management businesses) are reluctant to provide their raw data directly due to concerns about data security and resources [35, 39, 40]. In this situation, QoS data may become private and isolated, making current LFT techniques with the centralized modeling principle fail to operate. Consequently, a hot yet thorny issue arises, i.e., how to implement temporal pattern-aware QoS prediction with privacy-preserving.

An efficient method for protecting privacy in big-data applications is federated learning (FL) [35, 39, 41]. FL constructs a shared model on isolated data, integrating the desired information from all dispersed users' data into a shared model while protecting their privacy [42-45]. However, the majority of FL-based approaches cannot capture the dynamic patterns from QoS since they are constructed for static circumstances. How to predict accurately unknown QoS data with privacy-preserving remains a critical challenge.

To enable a temporal pattern-aware QoS predictor to protect privacy and extract temporal characteristics from QoS, this research innovatively designs a federated latent factorization of tensors (FL-LFT) model. FL-LFT has two-fold ideas: 1) it naturally takes into account the temporal factors by representing dynamic QoS data as tensors, and 2) a federated learning framework is built to make distributed users collaboratively train a shared LFT model while protecting user privacy. As a result, it is guaranteed to acquire a temporal pattern-aware QoS-predictor with privacy-preserving. This research has two contributions:

- An innovative FL-LFT model. It implements temporal pattern-aware QoS prediction with privacy-preserving by FL.
- An algorithm is designed and its corresponding time complexity is analyzed for the proposed FL-LFT model.

Extensive experiments on a QoS dataset are done to evaluate FL-LFT in scenes of privacy-preserving. The results demonstrate that the FL-LFT model's prediction accuracy is notably better than the state-of-the-art FL methods. Sections I-VI respectively are Introduction, Related Work, Preliminary, Methodology, Experiment, and Conclusion.

## RELATED WORK

### A. Latent Factorization of Tensors

Training three low-rank latent feature matrices to formulate approximations for the known values of QoS is the fundamental idea behind LFT [8, 27, 32, 46]. Unlike latent feature analysis models (LFA) [16, 20, 47-53] that separately handle time-oriented slices one by one, LFT is modeled on the whole QoS data through a three-order tensor. Hence, LFT can more effectively describe the dynamic characteristics of QoS. Some respective LFT models have been proposed to address temporally dynamical data [54, 55]. For instance, a WSPred framework for QoS prediction is proposed by *Zhang et al.* [29]. It employs stochastic gradient descent (SGD) [56-58] for training an LFT model. *Zhang et al.* [10] present a non-negative tensor factorization (NTF) model that can cope with an HDI tensor when subjected to non-negativity restrictions [59]. To describe the temporal variations of user-perspective QoS data, *Luo et al.* [16] present a biased non-negative LFT model.

Generally, when it involves capturing temporal characteristics from QoS, an LFT model is frequently extremely successful. However, all the above LFT models are centralized. They cannot protect user's privacy, which is unacceptable for increasingly privacy-sensitive users. In comparison, the presented FL-LFT model adopts the FL framework, which can not only accurately capture the dynamic features of QoS but also well protect the user's privacy in predicting unknown QoS data.

### B. Federated learning

The initial notion of FL arose from Google, which intended to develop a high-performance shared model through the utilization of isolated data across different clients [41, 60, 61]. A typical FL model is decentralized modeling combined with the approaches of encryption [62-64], differential privacy [65-67], feature/parameter extraction [68-70], etc., to preserve privacy. FL has been applied in various real applications for privacy-preserving [39, 71-74], such as Google's Gboard [41].

Currently, plenty of FL approaches have been presented for privacy-preserving collaborative data analysis [68], where LFA-based ones are the most representative [24, 35, 47, 62, 75]. For instance, A federated recommendation system named FedRec was put forward by *Lin et al.* [76] based on LFA. Later, *Liang et al.* improved FedRec by designating particular denoising clients with a privacy-aware strategy to decrease the impact of noise data [77]. *Lin et al.* [78] implement federated recommender systems by integrating meta-learning into LFA. Besides, additional methods were investigated to enhance FL's performance. For example, graph neural networks [1, 79, 80] were utilized to capture high-order information in FL-based recommender systems. Tensor decomposition is utilized to integrate more information into the modeling of FL [68, 81].

However, since the majority of the aforementioned FL techniques are designed for static scenarios, they cannot catch the dynamic feature of QoS. Although there are a few models based on tensor decomposition that could capture the dynamic properties of QoS, they are incompatible with the sparse properties of QoS since they are not data-density-oriented. In comparison, the FL-LFT model is designed for dynamic scenarios and is data-density-oriented, which guarantees its accurate and efficient prediction of QoS data with large-scale, dynamic, and sparse properties.

## PRELIMINARY

### A. Symbols and Notations

As illustrated in Fig.1, this research takes a user-service-time tensor as a fundamental research object. To clarify, Table I gives a summary of the primary symbols and their meanings.

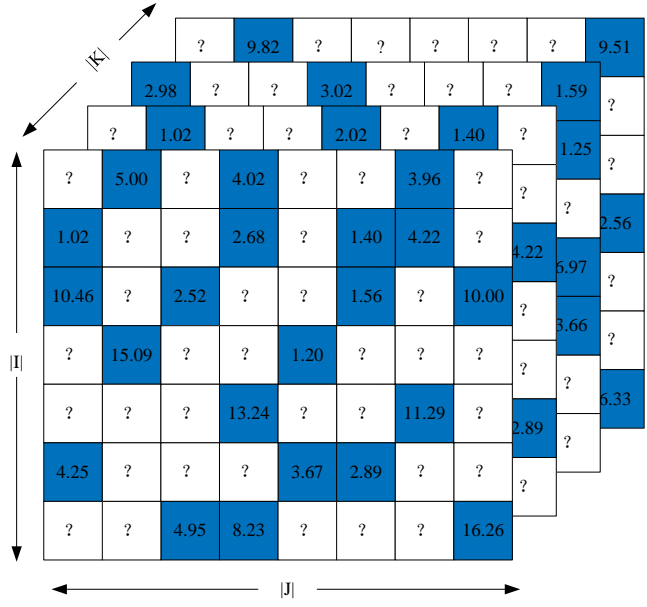
Fig.1. QoS tensor of user-service-time showing response time.

TABLE I. Symbols and Notations.

| Symbol | Explanation |
|---|---|
| **Y** | A third-order user-service-time tensor. |
| **Ŷ** | Low-rank approximation to **Y**. |
| **Y**$_r$ | Rank-one tensor regarding **Y**. |
| I, J, K | Sets corresponding users, services, and time points. |
| $y_{ijk}$, $y_{ijk}^{(r)}$ | A single element in **Y** and **Y**$_r$, respectively. |
| R | The rank of Ŷ and the dimension of latent feature space. |
| D, E, T | Latent feature matrices w.r.t. I, J, and K, respectively. |
| **d**$_i$, **e**$_j$, **t**$_k$ | The $i$-th, $j$-th, and $t$-th row-vector of latent matrix D for *the* $i$-th user, E and T. |
| **d**$_{.r}$, **e**$_{.r}$, **t**$_{.r}$ | Latent feature vectors w.r.t. the $r$-th rank-one tensor in Ŷ; also denote the $r$-th column-vector of D, E, and T. |
| $d_{ir}$, $e_{jr}$, $t_{kr}$ | Single latent feature in **d**$_i$, **e**$_j$, **t**$_k$. |
| Λ | Observed element set of **Y**. |
| Γ | Unobserved element set of **Y**. |
| $U_i$, U | The data set of the $i$-th user and all data from different users. |
| n | Training iteration count. |
| λ | The regularization coefficient of latent feature matrices. |

***Definition 1*** (***HDI User-Service-Time Tensor***). Given I, J, and K, $\mathbf{Y}^{|I|\times|J|\times|K|}$ is a user-service-time tensor. Its element $y_{ijk}$ indicates the QoS values observed by user $i \in I$ on service $j \in J$ at time point $k \in K$. Given that Λ and Γ represent **Y**'s observed element set and unobserved one, if $|\Lambda| \ll |\Gamma|$, **Y** is HDI.

*B. Latent Factorization of Tensors*

Given an HDI user-service-time tensor, an LFT model is effective for dynamic QoS analysis. As previously stated by the Canonical Polyadic decomposition principle [8, 27, 82], the target of an LFT model is to obtain its rank-$R$ approximation $\hat{\mathbf{Y}}^{|I|\times|J|\times|K|}$. **Ŷ** is expressed as a sum of $R$ rank-one tensors of **Y**$_1$, **Y**$_2$, …, **Y**$_R$ as follows:

$$\mathbf{Y} = \sum_{r=1}^{R} \mathbf{Y}_r, \qquad (1)$$

where $R$ represents the rank of the resultant approximation **Ŷ** to **Y** and is also the dimension of latent feature space.

***Definition 2*** (***Rank-one tenor***). If $\mathbf{Y}_r^{|I|\times|J|\times|K|}$ can be represented as the outer product of three latent feature vectors, it is a rank-one tensor. The expression is as follows:

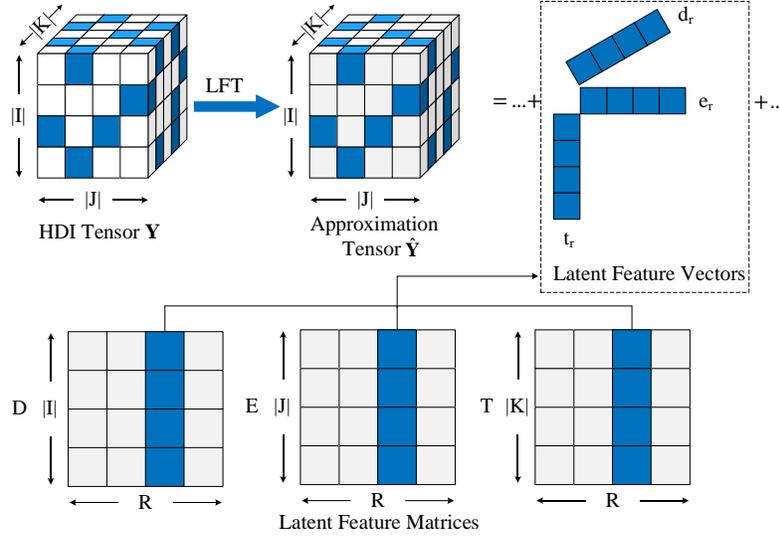

Fig.2. An LFT model following the Canonical Polyadic tensor factorization.

$$\mathbf{Y}_r = \mathbf{d}_{\cdot r} \circ \mathbf{e}_{\cdot r} \circ \mathbf{t}_{\cdot r}, \quad (2)$$

where 'o' denotes the operation of the outer product, and vector $\mathbf{d}_{\cdot r} \in R^{|I|}$, $\mathbf{e}_{\cdot r} \in R^{|J|}$, $\mathbf{t}_{\cdot r} \in R^{|K|}$ with $r=1, 2, \ldots, R$.

Fig. 2 illustrates that the rank-one ingredients $\mathbf{Y}_r$ can be acquired based on the $\mathbf{d}_{\cdot r}, \mathbf{e}_{\cdot r}, \mathbf{t}_{\cdot r}$ from $D^{|I| \times R}$, $E^{|J| \times R}$, $T^{|K| \times R}$. Therefore, the expected rank-one tensors are dependent on D, E, T. According to definition 2, any element $y_{ijk}^{(r)}$ of $\mathbf{Y}_r$ is represented as follows:

$$y_{ijk}^{(r)} = d_{ir} e_{jr} t_{kr}. \quad (3)$$

Therefore, every element $y_{ijk}$ of $\hat{\mathbf{Y}}$ is reformulated into

$$\hat{y}_{ijk} = \sum_{r=1}^{R} d_{ir} e_{jr} t_{kr}. \quad (4)$$

To acquire the intended D, E, and T, a loss function to differentiate $\hat{\mathbf{Y}}$ and $\mathbf{Y}$ is demanded. According to previous investigations [27, 83], the data density-oriented strategy effectively lowers both the storage and computational complexity of an LFT model. As a result, a loss function for this research is solely defined on the observed element set Λ of the target tensor. Hence, Euclidean distance is subsequently utilized [84-87] to formulate the loss function for an LFT model as follows:

$$\min_{\hat{\mathbf{Y}}} \varepsilon = \|\mathbf{Y} - \hat{\mathbf{Y}}\|_F^2 = \sum_{y_{ijk} \in \Lambda} \left(y_{ijk} - \hat{y}_{ijk}\right)^2 = \sum_{y_{ijk} \in \Lambda} \left(y_{ijk} - \sum_{r=1}^{R} d_{ir} e_{jr} t_{kr}\right)^2. \quad (5)$$

According to the previous study [88, 89], regularization should be incorporated into LFT to enhance its generalization capabilities. Therefore, the objective function can be formulated as follows:

$$\min_{\hat{\mathbf{Y}}} \varepsilon = \sum_{y_{ijk} \in \Lambda} \left(y_{ijk} - \hat{y}_{ijk}\right)^2 + \lambda \sum_{r=1}^{R} \left(d_{ir}^2 + e_{jr}^2 + t_{kr}^2\right). \quad (6)$$

By utilizing an appropriate optimization algorithm [8, 90] to minimize (6) for training D, E, and T, a well-learned LFT model can be acquired for accurately predicting unknown QoS elements.

## THE PROPOSED FL-LFT MODEL

### A. The Framework of FL-LFT

FL-LFT can cooperatively train a shared LFT model without requesting users to share or gather their original data in a central location. Thus, it avoids the leakage of any user's

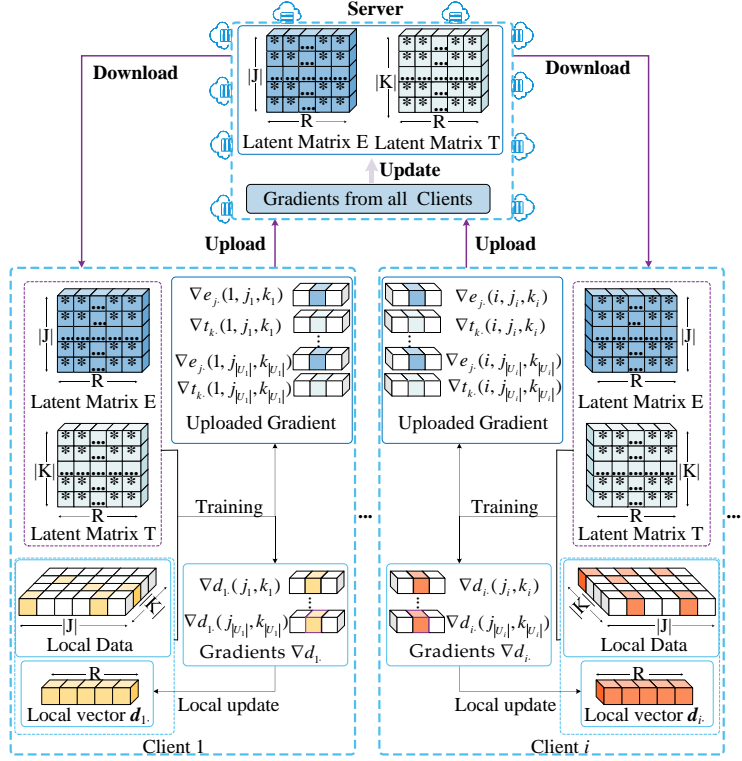

Fig.3. The overall framework of FL-LFT.

raw data. According to FedAvg [41], the loss function of FL-LFT is formulated as:

$$\min_{\hat{Y}} \varepsilon = \sum_{i \in I} \frac{|\Lambda_i|}{|\Lambda|} \sum_{y_{ijk} \in \Lambda_i} \left( y_{ijk} - \hat{y}_{ijk} \right)^2 + \lambda \sum_{r=1}^{R} \left( d_{ir}^2 + e_{jr}^2 + t_{kr}^2 \right). \quad (7)$$

The overall FL-LFT learning framework is shown in Fig.3. For each client, it stores the latent feature vector $d_i$ locally and maintains its data locally during the learning process. The server stores the latent feature matrices E and T which are shared by all clients.

As illustrated in Fig. 3, the interaction per round between the server and each client comprises the following steps:

- At the beginning, the server randomly utilizes small values to initialize the model parameters, i.e., E and T;
- Each client downloads the model parameters, i.e., E and T, from the server. Then, client $i$ utilizes its data and the downloaded model parameters to train its latent feature vector $d_i$;
- Client $i$ calculates the gradients, i.e., $\nabla e_{jr}(i, j, k)$ and $\nabla t_{kr}(i, j, k)$ for $r=(1, \ldots, R)$, locally. Then it uploads the gradients to the server;
- The server updates E and T after receiving gradients.

Note that the server updates E and T based on the gradients received from all the clients.

### B. Federated Optimization

Under the FL setting, the objective function in each client is reformulated as follows:

$$\min_{d_i} \varepsilon_{local}^i = \sum_{y_{ijk} \in \Lambda_i} \left( y_{ijk} - \hat{y}_{ijk} \right)^2 + \lambda \sum_{r=1}^{R} \left( d_{ir}^2 + e_{jr}^2 + t_{kr}^2 \right), \quad (8)$$

where $\Lambda_i$ indicates the collection of the user's known entry.

Our target is to acquire the latent feature vector $d_i$, along with the desired E and T. To accomplish this purpose, the alternative stochastic gradient descent approach[91] is utilized to solve (8).

**1) Federated optimization in the client**: Each client $i$ downloads latent feature matrix E and T from the server. Considering the instant loss $\varepsilon_{ijk}$ on a single element $y_{ijk}$ in (8), the following formula is acquired:

$$\min_{d_i} \varepsilon_{ijk} = \left( y_{ijk} - \hat{y}_{ijk} \right)^2 + \lambda \sum_{r=1}^{R} \left( d_{ir}^2 + e_{jr}^2 + t_{kr}^2 \right). \quad (9)$$

Then, the gradient $\nabla d_{ir}$ can be computed and the latent feature vector $d_i$ can be updated locally. Therefore, the latent vector $d_i$ can be updated as follows:

$$on\ y_{ijk},\ for\ r = 1 \sim R : d_{ir} \leftarrow d_{ir} - \eta \nabla d_{ir}, \quad (10)$$

where $\eta$ is the learning rate, and $\nabla d_{ir}$ is calculated as

$$\nabla d_{ir} = \left( y_{ijk} - \hat{y}_{ijk} \right)\left( -e_{jr} t_{kr} \right) + \lambda d_{ir}. \quad (11)$$

The gradients, i.e., $\nabla e_{jr}(i, j, k)$ and $\nabla t_{kr}(i, j, k)$, $x_{ijk} \in \Lambda_i$, can be calculated locally with the user's data, the latent feature vector $d_i$, and the downloaded model parameters as follows:

$$on\ y_{ijk},\ for\ r = 1 \sim R:$$
$$\begin{cases} \nabla e_{jr}(i, j, k) = \dfrac{\partial \varepsilon_{ijk}}{\partial e_{jr}} = \left( y_{ijk} - \hat{y}_{ijk} \right)\left( -d_{ir} t_{kr} \right) + \lambda e_{jr} \\ \nabla t_{kr}(i, j, k) = \dfrac{\partial \varepsilon_{ijk}}{\partial t_{kr}} = \left( y_{ijk} - \hat{y}_{ijk} \right)\left( -d_{ir} e_{jr} \right) + \lambda t_{kr} \end{cases}, \quad (12)$$

which are then uploaded to the server. The learning process of each client is detailed in Algorithm 2.

**2) Federated optimization in the server:** this part will detail how to update E and T by utilizing the gradients from all the clients in the server. Specifically, the server's loss function can be expressed as:

$$\min_{E,T} \varepsilon = \sum_{i \in I} \dfrac{|\Lambda_i|}{|\Lambda|} \sum_{y_{ijk} \in \Lambda_i} \left( y_{ijk} - \hat{y}_{ijk} \right)^2 + \lambda \sum_{r=1}^{R} \left( d_{ir}^2 + e_{jr}^2 + t_{kr}^2 \right). \quad (13)$$

Similarly, considering the instant loss $\varepsilon_{ijk}$ on a single element $y_{ijk}$ in (14), the following formula is obtained:

$$\min_{E,T} \varepsilon_{ijk} = \left( y_{ijk} - \hat{y}_{ijk} \right)^2 + \lambda \sum_{r=1}^{R} \left( d_{ir}^2 + e_{jr}^2 + t_{kr}^2 \right). \quad (14)$$

Hence, the E and T can be updated as follows:

$$on\ y_{ijk},\ for\ r = 1 \sim R : \begin{cases} e_{jr} \leftarrow e_{jr} - \eta \nabla e_{jr}(i, j, k) \\ t_{kr} \leftarrow t_{kr} - \eta \nabla t_{kr}(i, j, k) \end{cases}, \quad (15)$$

where the gradients, i.e., $\nabla e_{jr}(i, j, k)$, $\nabla t_{kr}(i, j, k)$, are received from all participants. The training process in the server is detailed in Algorithm 1.

*C. Complexity Analysis*

Each client's worst time complexity is $|\Lambda_i| \times (R+1) \times 4$ at each iteration. The server's worst time complexity is about $|I| \times |\Lambda_i| \times R \times 2$ at each iteration. Since $R$ and $|I|$ are usually fixed, each client $i$'s and the server's time complexity are linear to the number of $\Lambda_i$.

ALGORITHM 1: FL-LFT IN SERVER.

**Algorithm 1.** The algorithm of alternative stochastic gradient descent in the server perspective.

| | |
|---|---|
| 1 | Initialize the model parameters E and T with small random values, and set the maximum number of training iterations $N$. |
| 2 | while $n \leq N$ && not convergence |
| 3 |   For $i = 1, 2, …, |I|$ do parallelly |
| 4 |     Client (E, T, $\lambda$, $\eta$, $i$, $n$) |
| 5 |     For $y_{jk} \in \Lambda_i$ do |
| 6 |       Update E and T via (15). |
| 7 |     end for |
| 8 |   end for |

ALGORITHM 2: FL-LFT IN CLIENT.

**Algorithm 2.** The algorithm of alternative stochastic gradient descent in the client perspective.

| | |
|---|---|
| 1 | Initialization and download model parameters from the server. |
| 2 | For element $y_{ijk} \in \Lambda_i$ do |
| 2 |   Calculate the gradients, i.e., $\nabla d_{ir}$, and update $d_i$ via (10) and (11). |
| 4 |   Calculate the gradients, i.e., $\nabla e_{jr}(i, j, k)$, $\nabla t_{kr}(i, j, k)$, via (12). |
| 5 | end for |
| 6 | Upload the gradients, i.e., $\nabla e_{jr}(i, j, k)$, $\nabla t_{kr}(i, j, k)$, to the server. |

For the communication cost, the primary communication is that each client $i$ uploads $\nabla e_{jr}(i, j, k)$ and $\nabla t_{kr}(i, j, k)$ to the server and downloads the latent feature matrix E and T from the server. The gradients total $2 \times 8 \times R \times |\Lambda_i|$ byte information and the latent matrices from the server include $8 \times R \times (|J|+|K|)$ byte information. Thus, each client communication cost at each round is $8 \times R \times (|J|+|K|+2 \times |\Lambda_i|)$. The communication complexity is linear to the element amount of $\Lambda_i$. The server's communication complexity is $|I| \times 8 \times R \times (|J|+|K|)+16 \times |I| \times R \times |\Lambda_i|$, which is linear to the entry amount of $\Lambda$.

## EXPERIMENTS

The following experiments seek to explore if FL-LFT performs better in forecasting unknown QoS data than both state-of-the-start federated and non-federated models.

### A. General settings

**Datasets**. The experiments are conducted on a dataset collected by WSMonitor [29], whose details are summarized in Table II. In the experiments, each model will be trained with 5%-20% of the dataset and predict the remaining 95%-80% of data for evaluating their performance.

TABLE II. DATASET DETAILS.

| Dataset | Response-time(D1) |
|---|---|
| Scale | 0-20s |
| User Count | 142 |
| Service Count | 4500 |
| Time Point Count | 64 |
| Element Count | 30287611 |

**Evaluation Metrics**. The accuracy of predictions is the primary purpose of this work since it indicates whether or not the model has accurately captured the fundamental properties of an HDI tensor. Thus, the mean absolute error (MAE) [92] and root mean square error (RMSE) [17] are adopted as the metrics:

$$RMSE = \sqrt{\sum_{y_{ijk} \in \psi} (y_{ijk} - \hat{y}_{ijk})^2 / |\psi|}, MAE = \sum_{y_{ijk} \in \psi} |y_{ijk} - \hat{y}_{ijk}| / |\psi|,$$

where $\psi$ represents the test set. A higher forecast accuracy is shown by lower values of the MAE and RMSE.

**Baselines.** FL-LFE is compared with seven state-of-the-art models which are detailed in Table III.

TABLE III. SUMMARY OF ALL THE COMPARED MODELS.

| Model | Description |
|---|---|
| MF [93] | The LFA-based model. (IEEE Computer 2009). |
| GLocal_K [94] | The Global Local Kernel-based matrix completion model (CIKM 2021). |
| FPMF [76] | The federated recommender system for rating prediction is based on the probabilistic matrix factorization (IEEE Intelligent System 2021). |
| FSVD++ [76] | The federated multifaceted collaborative filtering model with privacy-preserving (IEEE Intelligent System 2021). |
| MetaMF [78] | The neural federated collaborative ranking is based on memory-based attention (SIGIR 2020). |
| FedRec++ [77] | A federated recommender without loss depends on assigning denoise clients to decrease the impact of noisy data with the privacy-aware method. (AAAI 2021). |
| Light FR [35] | A lightweight federated matrix factorization method for the privacy-preserving recommendation. (TOIS 2023) |
| FL-LFT | The proposed framework in this work. |

TABLE IV. THE RESULTS OF PREDICTION ACCURACY COMPARISON.

| Dataset Case | Metric | Federated | | | | | Non-federated | | Our model |
|---|---|---|---|---|---|---|---|---|---|
| | | *FPMF* | *FSVD++* | *MetaMF* | *FedRec++* | *LightFR* | *MF* | *Glocal_K* | *FL-LFT* |
| D1.1(5%) | RMSE | 3.6782 | 3.6555 | 3.6541 | 4.1299 | 4.2051 | 3.7958 | 9.0357 | 2.8856 |
| | MAE | 1.701 | 1.6506 | 1.6864 | 1.9882 | 2.5226 | 1.7361 | 6.5737 | 1.3421 |
| D1.2(10%) | RMSE | 3.2981 | 3.2835 | 3.4319 | 3.7758 | 3.8395 | 3.3612 | 8.9471 | 2.7708 |
| | MAE | 1.4691 | 1.4591 | 1.4766 | 1.9882 | 2.2165 | 1.5441 | 6.5473 | 1.2807 |
| D1.3(15%) | RMSE | 3.1456 | 3.1212 | 3.4019 | 3.7489 | 3.8014 | 3.1663 | 8.9357 | 2.7508 |
| | MAE | 1.4036 | 1.3841 | 1.4038 | 1.9153 | 2.1003 | 1.4417 | 6.4901 | 1.2562 |
| D1.4(20%) | RMSE | 3.0591 | 3.0257 | 3.2801 | 3.7159 | 3.7715 | 3.0501 | 8.9356 | 2.7489 |
| | MAE | 1.3718 | 1.3448 | 1.3892 | 1.9018 | 1.9946 | 1.3791 | 6.4454 | 1.2476 |
| Statistics | Win/loss | 8/0 | 8/0 | 8/0 | 8/0 | 8/0 | 8/0 | 8/0 | **56/0** |
| | F-rank★ | 3.375 | 2.125 | 4.125 | 6.0 | 7.0 | 4.375 | 8 | **1.0** |
| | *p*-value◊ | **0.0039** | **0.0039** | **0.0039** | **0.0039** | **0.0039** | **0.0039** | **0.0039** | -- |

★A lower value represents a better performance. ◊ FL-LFT notably exceeds the related comparison model if the *p*-value is lower than 0.05.

**Implement details.** The hyperparameters of FL-LFT are set as follows: learning rate $\eta$=0.00038, regularization coefficient $\lambda$=0.001. Each case will go through five rounds of testing, and the average performance will be the last result.

*B. Prediction Accuracy Comparison*

The comparison reports are displayed in Table IV. Several statistical techniques [1, 95, 96] are applied to strengthen the analysis of these results, containing a win/loss comparison, the Wilcoxon signed-ranks test, and the Friedman test.

First, a comparison is performed between non-federated models and FL-LFT. As shown in Table IV, since MF and Glocal_K cannot capture dynamic patterns concealed in QoS data, they are significantly exceeded by the proposed FL-LFT model. For instance, MF's RMSE is 3.7958, which is about 23.97% higher than FL-LFT's 2.8856 on D1.1. MF's MAE is 1.7361, which is about 22.69% higher than FL-LFT's 1.3421 on D1.1. Similar conclusions can be obtained on the other cases between MF and FL-LFT.

On the other hand, it appears that FL-LFT outperforms all other federated models in terms of prediction accuracy. As shown in Table IV, FL-LFT's performance is notably better than five federated models in all involved cases. For instance, the prediction accuracy of FSVD++ which is the best among five federated models, is still exceeded by FL-LFT. FSVD++'s RMSE is 3.0257 on D1.4, which is 9.14% higher than FL-LFT's 2.7489. FSVD++'s MAE is 1.3448 which is about 7.22% higher than the 1.2496 achieved by FL-LFT. Similar conclusions can be obtained in other cases between FSVD++ and FL-LFT.

To sum up, these results validate that the presented FL-LFT model displays significantly superior prediction accuracy than the seven comparison models. The reason is that all compared models except for FL-LFT do not capture the temporal dynamics of an HDI tensor, which is crucial to predict unknown elements in QoS.

## CONCLUSION

To deal with the crucial problem of privacy-preserving QoS prediction, this paper presents an FL-LFT model that can capture the temporal dynamics of a user-service-time tensor. It notably exceeds existing FL models. However, the current FL-

LFT model fails to consider that these gradients may leak important data information if no security guarantee is provided. This problem will be addressed by us in the future.